\documentclass[letterpaper]{article} % DO NOT CHANGE THIS
\usepackage{aaai2026}  % DO NOT CHANGE THIS
\usepackage{times}  % DO NOT CHANGE THIS
\usepackage{helvet}  % DO NOT CHANGE THIS
\usepackage{courier}  % DO NOT CHANGE THIS
\usepackage[hyphens]{url}  % DO NOT CHANGE THIS
\usepackage{graphicx} % DO NOT CHANGE THIS
\usepackage{natbib}  % DO NOT CHANGE THIS AND DO NOT ADD ANY OPTIONS TO IT
\usepackage{caption} % DO NOT CHANGE THIS AND DO NOT ADD ANY OPTIONS TO IT
\usepackage{algorithm}
\usepackage{algorithmic}
\usepackage{amsmath}
\usepackage{amssymb}
\usepackage{booktabs}
\usepackage{pifont}
\usepackage{newfloat}
\usepackage{listings}
\DeclareCaptionStyle{ruled}{labelfont=normalfont,labelsep=colon,strut=off} % DO NOT CHANGE THIS
\floatstyle{ruled}
\newfloat{listing}{tb}{lst}{}
\floatname{listing}{Listing}
\title{Beyond Accuracy: Introducing a Symbolic-Mechanistic Approach to Interpretable Evaluation}
\author{
    Reza Habibi\textsuperscript{\rm 1}\equalcontrib,
  Darian Lee\textsuperscript{\rm 1}\equalcontrib,\\
   Magy Seif El-Nasr
}
\affiliations{
    \textsuperscript{\rm 1} University of California, Santa Cruz\\
   606 Engineering Loop, Santa Cruz, CA 95064 USA\\
    rehabibi@ucsc.edu, daeilee@ucsc.edu
}

\usepackage{bibentry}
\begin{document}

\maketitle

\begin{abstract}
Accuracy-based evaluation cannot reliably distinguish genuine generalization from shortcuts like memorization, leakage, or brittle heuristics, especially in small-data regimes. In this \textbf{position paper}, we argue for mechanism-aware evaluation that combines task-relevant symbolic rules with mechanistic interpretability, yielding algorithmic pass/fail scores that show exactly where models generalize versus exploit patterns. We demonstrate this on NL-to-SQL by training two identical architectures under different conditions: one without schema information (forcing memorization), one with schema (enabling grounding). Standard evaluation shows the memorization model achieves 94\% field-name accuracy on unseen data, falsely suggesting competence. Our symbolic-mechanistic evaluation reveals this model violates core schema generalization rules, a failure invisible to accuracy metrics.
\end{abstract}

% Uncomment the following to link to your code, datasets, an extended version or similar.
% You must keep this block between (not within) the abstract and the main body of the paper.
% \begin{links}
%     \link{Code}{https://aaai.org/example/code}
%     \link{Datasets}{https://aaai.org/example/datasets}
%     \link{Extended version}{https://aaai.org/example/extended-version}
% \end{links}

\section{Introduction and Background}
Standard NLP evaluation relies on held-out test sets with metrics like exact match, F1, or BLEU \cite{papineni2002bleu,nakache2005evaluation,gehrmann-etal-2023-repairing}. However, recent work reveals these methods cannot reliably distinguish memorization from generalization. Contamination is widespread, with benchmarks like QNLI showing over 50\% contamination in popular pretraining corpuses \cite{lee-etal-2022-deduplicating, dodge-etal-2021-documenting}, and models exploit spurious heuristics even on clean data \cite{gururangan-etal-2018-annotation, mccoy-etal-2019-right, kim-etal-2025-benchmark}. Alternative methods like LLM-as-a-judge suffer from self-preference bias and adversarial vulnerabilities \cite{chen-etal-2024-humans, raina-etal-2024-llm}. We argue these failures stem from a fundamental problem: current evaluation measures surface-level performance rather than underlying computational mechanisms. We propose mechanistic interpretability (MI) evaluation that directly examines whether models use correct algorithms.

\subsection{Contamination and Exploitable Heuristics}

Contamination research has demonstrated the scope of this issue. \cite{zhang2024careful} found 0.32 correlation between memorization probability and performance on GSM8K variants, while \cite{li2024task} found 0.88 correlation between exact-match generation without context and test performance. \cite{aiyappa-etal-2023-trust} observed ChatGPT's benchmark performance improved after papers evaluating it were published, indicating RLHF-driven contamination. Critically, \cite{samuel2025towards} found alarmingly low agreement between contamination detection methods, indicating no reliable standard exists, implying that mitigation and detection of contamination before testing is not a viable option. Additionally, modern training pipelines with RLHF on user data and continuous web scraping create ongoing contamination pathways that cannot be eliminated \cite{zhang2024careful,aiyappa-etal-2023-trust}.

Even with pristine data, models exploit spurious patterns. \cite{gururangan-etal-2018-annotation} achieved 67\% SNLI accuracy using only hypothesis sentences by exploiting annotation artifacts. \cite{kim-etal-2025-benchmark} showed mechanistically that benchmarks often fail to test their advertised skills, instead rewarding alternative mechanisms. For tasks with small datasets, common in low-resource languages and specialized domains, test sets lack statistical power to separate genuine competence from pattern matching.

\subsection{From Grokking to Verification}

Research on grokking demonstrates that generalization corresponds to distinct internal structure \cite{power2022grokking, nanda2023progress, lui_groking}. \cite{nanda2023progress} showed that generalization circuits form during memorization before full utilization, suggesting mechanistic probing can detect whether models operate in memorization- versus generalization-dominant regimes. However, grokking insights remain unapplied to real NLP evaluation.

This motivates symbolic verification. Symbolic AI provides frameworks for expressing correctness properties as semantically meaningful rules \cite{old_logic_paper, newell1959report, commonsense, xie2022neurosymbolic}. By checking whether internal computations match expected generalization patterns, symbolic rules enable more rigorous evaluation than accuracy alone.

\subsection{Proposal}

Our approach employs ``non-negotiable rules" in symbolic logic describing properties any circuit reliably solving the task must satisfy. Given task T, the user specifies rules R = \{r\_1, ..., r\_k\} capturing essential requirements at the level of task semantics: what information must be used, what invariances must hold. Rules must be verifiable via common mechanistic interventions like activation patching, logit lens analysis, or attention visualization.

For each example in a small evaluation set, we test whether rules are satisfied, assigning pass/fail scores. The aggregate percentage indicates how consistently the model uses proper generalization circuits. 

This paper is a \textit{position paper} advocating mechanism-aware evaluation for tasks with a well-defined intended algorithm (e.g., grounding, retrieval, or parsing). We do not claim applicability to open-ended or creative generation tasks, where multiple qualitatively different algorithms may be equally valid and no single mechanistic criterion is appropriate.

\section{Case Study}

We compare our evaluation framework to standard accuracy on a simple NL-to-SQL task, focusing on schema grounding as the core algorithm. We train two models with identical architectures but different training conditions: one without schema information (NO\_Schema), forcing reliance on shallow heuristics, and one with schema (Schema), enabling genuine generalization. We show that standard evaluation suggests roughly equal capabilities, whereas symbolic mechanistic evaluation reveals the NO\_Schema model's lack of true generalizability.
\subsubsection{Dataset}
Our case study employs the TinySQL CS1 Synonyms dataset \cite{harrasse-etal-2025-tinysql}. Each example consists of an English prompt (\texttt{english\_prompt}), a database schema provided as a \texttt{CREATE TABLE} statement (\texttt{create\_statement}), and the target SQL query (\texttt{sql\_statement}). Critically, the schema uses a synonym of the column name mentioned in the English prompt,\footnote{E.g., schema \texttt{CREATE TABLE backup (website TEXT)}, prompt ``Pull up url from safekeeping copy'', target \texttt{SELECT website FROM backup}.} ensuring that correct SQL cannot be generated from natural language alone without schema grounding.

The dataset contains only simple queries following the pattern \texttt{SELECT x FROM y}. Given this structural simplicity, we focus evaluation on schema grounding, the ability to correctly map English column names to their schema synonyms, as the core algorithmic capability. The remaining query structure follows a predictable template across all examples.

\subsubsection{Training Conditions:}We finetune two models based on TinyStories-33M \cite{eldan2023tinystories}, with identical architecture and optimization settings, following \cite{harrasse-etal-2025-tinysql}. The models differ only in schema presence during training. The baseline was pretrained solely on TinyStories-33M short stories with no NL-to-SQL examples, ensuring no task contamination \cite{eldan2023tinystories}.

The \textbf{schema} condition includes the \texttt{create\_statement} in context for every training example. The \textbf{NO\_Schema} condition omits it entirely while keeping the same target SQL, forcing reliance on shallow heuristics rather than schema consultation. The column and table names in the English prompt are synonyms rather than exact matches, thus models benefit from using schema information when available \cite{harrasse-etal-2025-tinysql}.

We use a single standardized prompt format: \texttt{\#\#\# Instruction: \{english\_prompt\} \#\#\# Context: \{create\_statement or empty\} \#\#\#Response: \{sql\_statement\}}. Decoding begins after "\#\#\# Response:", and predicted SQL is compared against \texttt{sql\_statement}.

\subsubsection{Standard Accuracy-based Testing:}
We compare our approach to standard evaluation on the TinySQL CS1 Synonyms test split, reporting exact match accuracy (predictions identical to gold SQL) and field name accuracy (correct table/column names as a ratio of total gold field names, excluding SQL keywords). Each model is evaluated under two configurations: schema (schema included in test prompt) and NO\_Schema (schema excluded).

\subsection{Symbolic Mechanistic Evaluation}

We evaluate 100 matched prompt pairs where each has a \emph{clean} prompt with the correct schema field and a \emph{corrupted} version with an alternative word, holding format constant.

We test five corruption types to probe schema grounding. \textbf{DB-synonyms} replace column names with semantically related alternatives that also appeared as column names in training (``price''/``cost''), testing whether the model maps semantic equivalents within the trained schema. \textbf{Non-DB synonyms} use semantically related words that were never column names (``mouse''/``rat''), testing whether this semantic flexibility generalizes. The remaining three test whether the model is sensitive to schema violations at all, measuring whether it can detect and react when an unexpected word appears. \textbf{DB-scramble} pairs unrelated words that both appeared as column names (``price''/``brand''), \textbf{Non-DB scramble} pairs unrelated words that were not column names (``mouse''/``car''), and \textbf{Super-scramble} mixes column names with non-column words (``price''/``rat''). If the model detects these scrambled corruptions, it demonstrates active tracking of the schema.

\subsubsection{Symbolic Rules}
To evaluate whether a schema-checking circuit exists, we adopt a symbolic rule-based approach that decomposes this question into three testable conditions. First, we test causal sensitivity: does corrupting the schema token change which answer the model prefers? Second, we test localization: can we recover this effect by patching activations at the schema token position? Third, we test consistency: do multiple examples recover through the same layer, suggesting a reusable mechanism? We call these rules R1 through R3. Because R3 must only pass if R1 and R2 pass, we denote that a pass on R3 is analogous to all conditions being met. 
We measure preference for the correct token using logit-difference $\text{logit\_diff}(x; i) = \ell(x)_{\text{correct}_i} - \ell(x)_{\text{incorrect}_i}$, where $\ell(x)$ denotes model logits at the final position. Let $x_i^{\text{clean}}$ and $x_i^{\text{corr}}$ denote clean and corrupted prompts differing only in the schema token. We define the schema-sensitivity gap as
\[
\Delta_i = \text{logit\_diff}(x_i^{\text{clean}}; i) - \text{logit\_diff}(x_i^{\text{corr}}; i).
\]
A positive $\Delta_i$ indicates the model causally depends on the schema token when selecting answers.

\subsubsection{Activation Patching Protocol}

To test whether $\Delta_i$ can be recovered from specific layers, we perform residual-stream activation patching at the schema token position ($p$). We cache activations from $x_i^{\text{clean}}$, then for each layer $L \in \{0, \ldots, L_{\max}\}$, re-run $x_i^{\text{corr}}$ while replacing the residual activation at $(L,p)$ with the cached clean activation, yielding $\text{shift}_i(L) = \text{patched\_diff}_i(L) - \text{corr\_diff}_i$.  We define the best signed recovery and the layer achieving it as
\[
\text{best\_recovery}_i = \max_{L}~\max\!\bigl(0,\; \operatorname{sign}(\Delta_i)\cdot \text{shift}_i(L)\bigr)
\]
\[
L^*_i = \operatorname{argmax}_{L}~\max\!\bigl(0,\; \operatorname{sign}(\Delta_i)\cdot \text{shift}_i(L)\bigr)
\]
We cap recovery at $|\Delta_i|$ and define fractional recovery as $\text{RecFrac}_i = \min(\text{best\_recovery}_i, |\Delta_i|) / |\Delta_i|$. We denote $\text{TopLayers}_i$ as the set of layers achieving $\ge$90\% recovery, recognizing potential superposition and redundant circuits \citep{elhage2022toy}.

\subsubsection{Hierarchical Rule System}
Intuitively, our evaluation asks three questions: (1) does changing the schema token causally affect the model’s answer preference, (2) can this effect be localized to a specific internal computation via patching, and (3) is the same computation reused across inputs? The following definitions formalize these questions using logit-difference measurements and residual-stream interventions.
Starting with a top-level claim about circuit existence, we decompose into per-example causal tests (R1, R2, R3). 

Let $\tau_{\text{gap}} \ge 0$ be a minimum gap threshold, and $\alpha \in (0,1]$ be the required recovered fraction.

\paragraph{R1 (schema\_sensitivity).} If the schema token is corrupted, the model's preference changes by at least $\tau_{\text{gap}}$:
\[
R1_i \;=\; \mathbb{I}\bigl[|\Delta_i| \ge \tau_{\text{gap}}\bigr].
\]
This tests causal dependence on the schema token. Examples with $|\Delta_i| < \tau_{\text{gap}}$ indicate the model ignores schema information. Passing R1 shows the model's answer preference depends on which schema token is present.

\paragraph{R2 (recovery\_efficacy).} If R1 holds and the clean schema token activation is patched at position $(p)$, then at least $\alpha$ fraction of the gap is recovered:
\[
R2_i \;=\; R1_i \;\wedge\; \mathbb{I}\bigl[\text{RecFrac}_i \ge \alpha\bigr].
\]
This tests whether the causal effect localizes to a specific layer. Measuring recovery as a fraction ($\text{RecFrac}_i$) assesses mechanism presence independently of gap magnitude. Passing R2 shows a single layer at the schema position is sufficient to recover preference, indicating a localized circuit. Due to superposition or redundant circuits, multiple layers can achieve $\ge$90\% recovery, captured in $\text{TopLayers}_i$.

\paragraph{R3 (circuit\_reusability).} If R1 and R2 hold and the most common high-recovery layer across all examples is $L^*$, then $L^* \in \text{TopLayers}_i$:
\[
R3_i \;=\; R1_i \;\wedge\; R2_i \;\wedge\; \mathbb{I}\bigl[L^* \in \text{TopLayers}_i\bigr].
\]
This tests whether recovery uses a consistent layer across examples. Passing R3 shows the model uses the same computational pathway for different queries, indicating reuse of a consistent computational pathway, as opposed to relying on input-specific heuristics.

% Critically, R2 evaluates recovery \emph{quality} as a fraction of the observed gap. A model that degrades from 95\% to 92\% accuracy (gap = 0.03) and recovers to 94.7\% (RecFrac = 0.90) satisfies R2 identically to a model degrading from 80\% to 50\% (gap = 0.30) and recovering to 77\% (RecFrac = 0.90). This ensures we measure mechanism presence rather than conflating it with task difficulty or corruption strength.

We run the identical evaluation pipeline on our schema-trained model and NO\_Schema baseline model, and compare per-example pass rates ($\frac{1}{N}\sum_i R3_i$) under the same prompt set and thresholds. Remember that R3 can only pass if R1 and R2 pass as well.

% \begin{table}[t]
% \centering
% \small
% \begin{tabular}{l p{0.68\columnwidth}}
% \toprule
% \textbf{Rule} & \textbf{Statement} \\
% \midrule
% $R_1$ & (\texttt{schema\_sensitivity}) Schema corruption changes preference by $\ge \tau_{\text{gap}}$: $|\Delta_i| \ge \tau_{\text{gap}}$, indicating causal dependence. \\[0.5em]
% $R_2$ & (\texttt{recovery\_efficacy}) If $R_1$ holds and $\text{RecFrac}_i \ge \alpha$, patching $p{=}18$ recovers $\ge \alpha$ of the gap, indicating localization. \\[0.5em]
% $R_3$ & (\texttt{circuit\_reusability}) If $R_1$ and $R_2$ hold and $L^* \in \text{TopLayers}_i$, recovery uses the most common layer $L^*$, indicating reusability. \\
% \bottomrule
% \end{tabular}
% \caption{Hierarchical rules test: (R1) causal sensitivity to schema corruption, (R2) localized recovery via patching, (R3) consistent layer usage across examples.}
% \label{tab:rules}
% \end{table}

This framework yields an example-level pass rate estimating proximity to complete generalization. Decoupling gap significance (R1) from recovery quality (R2) measures mechanism presence independently of task difficulty or corruption strength. The consistency metric (R3) distinguishes position-specific effects from genuine circuit reuse, providing stronger evidence for localized computation.

\section{Results}

\subsection{Standard Accuracy}
\label{sec:standard_accuracy}
The TinySQL CS1 synonyms test-set results reveal a critical limitation of surface-level evaluation, as the model trained without schema achieves 93.5\% table/column accuracy even when schema is withheld at test time. Since the model has no access to schema information, this high performance can only result from exploiting spurious patterns in the data. Although the schema-trained model performs better (99.1\% with schema present), these metrics alone cannot distinguish whether the model genuinely uses schema information or simply memorizes training patterns (Table \ref{fig:eval_results}).

\newcommand{\cmark}{\ding{51}}
\newcommand{\xmark}{\ding{55}}

\begin{table}[h!]
\centering
\normalsize
\begin{tabular}{cccc}
\toprule
\textbf{Train} & \textbf{Eval} & \textbf{Exact} & \textbf{Field} \\
\textbf{Schema} & \textbf{Schema} & \textbf{Match} & \textbf{Acc} \\
\midrule
\xmark & \xmark & \textbf{72.1\%} & \textbf{93.5\%} \\
\xmark & \cmark & 10.4\% & 83.6\% \\
\cmark & \xmark & 0.0\% & 40.4\% \\
\cmark & \cmark & \textbf{94.0\%} & \textbf{99.1\%} \\
\bottomrule
\end{tabular}
\caption{Standard accuracy across model configurations. The NO\_Schema model achieves 93.5\% field-name accuracy by exploiting patterns, demonstrating traditional evaluation's unreliability.}
\label{fig:eval_results}
\end{table}

\subsection{Symbolic Mechanistic Results}
%this is with rabndom seed 12 in all places
On our 100 controlled prompt pairs, the schema-trained model exhibited significantly larger sensitivity to schema corruption (mean $\Delta_{\text{schema}} = 1.88$ [95\% CI: 1.44, 2.32]) compared to the NO\_Schema baseline (mean $\Delta_{\text{NO\_Schema}} = 0.65$ [95\% CI: 0.39, 0.90]), yielding a training effect of $+1.23$ [95\% CI: 0.79, 1.67] logits (190\% increase). Per-category analysis (Table~\ref{tab:category_results}) shows the effect is largest for scrambled corruptions (Super-Scramble: $+2.35$, Non-DB-Scramble: $+1.61$) and smaller for synonyms ($+0.53$ to $+0.57$), suggesting the model partially accepts semantically related substitutions.

Applying our symbolic rule framework (R1: $\Delta > 0.4$; R2: recovery $\geq 90\%$ of $|\Delta|$; and R3: $L^* \in \text{TopLayers}_i$)  \textbf{the schema model achieved 76\% overall PASS rate (76/100 examples) versus 59\% for the NO\_Schema model}\footnote{Alternative thresholds (R1: $\Delta \in \{0.25, 0.3\}$; R2/R3: $\geq 80\%, 95\%$) showed schema model outperforming baseline by 10-20pp across all combinations, suggesting rules are robust to threshold noise.}(Table~\ref{tab:category_results}). Category-specific PASS rates ranged from 55\% (DB-Synonyms) to 95\% (Non-DB-Scramble) for the schema model, consistently exceeding the NO\_Schema baseline across all categories. The consistent advantage across all five categories and large effect sizes with non-overlapping CIs support the framework's discriminative power.

Passing examples exhibited strong circuit consistency: among 76 schema-model examples satisfying R3 (indicating all three rules passed), Layers 0--2 collectively accounting for 89\%. In contrast, the NO\_Schema model showed distributed processing (modal layer usage: 34\%). This layer-level convergence in combination with high patching recovery rates provides evidence for a localized, reusable schema-checking circuit in the schema-trained model.

% Average recovery metrics confirm robust causal mediation: schema-model passing examples recovered mean $2.20\times$ the gap (range: 1.00--4.32), with mean best recovery of $+0.98$ logits. Non-DB-Scramble achieved the strongest recovery ($3.71$ logits, 95\% PASS), while DB-Synonyms showed weakest recovery ($0.61$ logits, 60\% PASS).

\begin{table}[h]
\small
\centering
\begin{tabular}{lccc}
\toprule
\textbf{Category} & \textbf{Effect ($\Delta_S - \Delta_N$)} & \textbf{PASS$_S$} & \textbf{PASS$_N$} \\
\midrule
DB-Synonyms & +0.53 & 55\% & 25\% \\
DB-Scramble & +0.51 & 75\% & 70\% \\
Non-DB-Syn & +0.57 & 80\% & 45\% \\
Non-DB-Scr & +2.28 & 95\% & 85\% \\
Super-Scr & +0.77 & 75\% & 70\% \\
\midrule
\textbf{Overall} & \textbf{+0.93} & \textbf{76\%} & \textbf{59\%} \\
\bottomrule
\end{tabular}
\caption{Per-category results across 100 examples (20 per category).}
\label{tab:category_results}
\end{table}

\section{Discussion}

Our results show that mechanistic evaluation distinguishes algorithmic learning from pattern-matching more effectively than standard metrics. The schema model achieved 76\% pass rate while the NO\_Schema baseline reached 59\%, a 17-percentage-point gap (compared to a 5-percentage-point gap in standard accuracy testing). This inversion shows that surface metrics dramatically underestimate the difference between genuine algorithmic learning and pattern exploitation. 

The symbolic mechanistic evaluation distinguishes these cases where standard metrics fail. The symbolic rule system provides interpretable diagnostics where R1 failures indicate the model ignores critical input features, R2 failures reveal distributed rather than localized computation, and R3 failures show inconsistent processing across examples. 

Current benchmarks face widespread data leakage with no reliable detection consensus. Models exploit spurious patterns even in clean test sets. Our NO\_Schema baseline demonstrates this by achieving 93.5\% field-name accuracy despite lacking the intended algorithm. Standard evaluation cannot detect this failure mode, however our symbolic-mechanistic approach correctly identifies this limitation.

\section{Conclusion and Future Work}
% This work introduces a mechanism-aware evaluation framework combining symbolic rules with mechanistic interpretability to distinguish genuine generalization from pattern exploitation. Applied to NL-to-SQL, our approach reveals that a model achieving 93.5\% field-name accuracy achieves only a 59\% PASS rate on mechanism consistency checks, demonstrating that standard metrics cannot reliably assess whether models use intended algorithms. The hierarchical rule system provides interpretable diagnostics, offering a template for tasks with well-defined algorithmic primitives such as retrieval, grounding, or structured parsing.

This work introduces a mechanism-aware evaluation framework combining symbolic rules with mechanistic interpretability to separate genuine generalization from pattern exploitation. Applied to NL-to-SQL, we show that a model reaching 93.5\% field-name accuracy achieves only 59\% on mechanism consistency checks, demonstrating that standard metrics cannot reliably assess whether models use intended algorithms. The hierarchical rule system provides interpretable diagnostics and serves as a template for tasks with well-defined algorithmic primitives such as retrieval, grounding, or structured parsing.

Three limitations point to productive future directions. First, our rules depend on hyperparameters requiring principled calibration; control tasks with known ground-truth mechanisms are needed for systematic threshold selection across domains. Second, activation patching risks distribution shifts that may allow recovery through confounding pathways; while we use convergent validation across multiple intervention types, future work should triangulate with path patching, causal tracing, or ablation studies. Third, symbolic rules become hard to define for complex tasks lacking clear algorithmic primitives, our framework suits targeted capability assessment more than holistic evaluation.

More broadly, mechanism-aware evaluation surfaces failure modes invisible to accuracy metrics. As model contamination grows endemic and pattern matching grows more sophisticated, verifying that models implement intended algorithms—not merely approximate correct output becomes essential. Future benchmarks should require mechanism certification alongside accuracy reporting, especially in high-stakes domains where reasoning transparency is critical.

% Key limitations suggest productive directions for future research. First, our rules depend on hyperparameters selected to balance sensitivity and specificity. Principled threshold selection requires control tasks with known ground-truth mechanisms for systematic calibration across domains.

% Second, activation patching can create distribution shifts that may allow recovery through confounding pathways. While we mitigate this through convergent validation across multiple intervention types, single-method causal claims remain provisional. Future work should triangulate findings using complementary techniques such as path patching, causal tracing, or ablation studies to strengthen mechanistic inferences.

% Third, defining symbolic rules becomes difficult for complex tasks lacking clear algorithmic primitives. While schema grounding admits straightforward verification, open-ended generation may permit multiple valid mechanisms, making it unclear what to certify. Our framework is best suited for targeted capability assessment rather than holistic evaluation.

% More broadly, this work demonstrates that mechanism-aware evaluation can detect failure modes invisible to accuracy metrics. As contamination becomes endemic and models grow capable of sophisticated pattern matching, verification that models implement intended algorithms rather than approximate correct outputs becomes essential. Future benchmarks should require mechanism certification alongside accuracy reporting, particularly for high-stakes domains where reasoning transparency is critical.

\bibliography{aaai2026}

\end{document}